\documentclass[letterpaper, 10 pt, conference]{ieeeconf} 

\usepackage{graphicx}

\IEEEoverridecommandlockouts               

\title{\LARGE \bf
Analysis of Social Robotic Navigation approaches: CNN Encoder and Incremental Learning as an alternative to Deep Reinforcement Learning
}

\author{Janderson Ferreira$^{1}$, Agostinho A. F. J\'unior$^{1}$, Let\'icia Castro$^{1}$, Yves M. Galv\~ao$^{1}$,
\\
Pablo Barros$^{2}$ and 
Bruno J. T. Fernandes$^{1}$.

\thanks{This study was financed in part by the Coordena\c{c}\~ao de Aperfei\c{c}oamento de Pessoal de N\'ivel Superior - Brasil (CAPES) - Finance Code 001, and the Brazilian agencies FACEPE and CNPq.}

\thanks{$^{1}$Authors are in the Programa de Engenharia da Computação, Escola Politécnica da Universidade de Pernambuco, POLI-UPE, Recife, Brazil.
{\tt\small \{jrb,aafj,lcpo,ymg,bjtf\}@ecomp.poli.br}}
 
\thanks{$^{2}$Pablo is part of the Cognitive Architecture for Collaborative Technologies (CONTACT) Unit - Istituto Italiano di Tecnologia, Genova, Italy.
{\tt\small pablo.alvesdebarros@iit.it}}
}

\begin{document}
\maketitle



\begin{abstract}

Dealing with social tasks in robotic scenarios is difficult, as having humans in the learning loop is incompatible with most of the state-of-the-art machine learning algorithms. This is the case when exploring Incremental learning models, in particular the ones involving reinforcement learning. In this work, we discuss this problem and possible solutions by analysing a previous study on adaptive convolutional encoders for a social navigation task.

\end{abstract}

\section{INTRODUCTION}

Path planning algorithms first emerged in 1956, proposed by Dijkstra \cite{dijkstra}, and since then have had several different applications. They have been instrumental for advancement in the areas of motion planning for robots, finding the safest path for pedestrians, autonomous vehicles, and personal mobility assistance \cite{mobilityassistance,mobilityassistance2,jds,jds2}. However, path planning presents high computational costs in large and dynamic environments. Also, to insert robots in daily human life activities, they should respect social norms of locomotion. For instance, they should avoid moving between two people walking together and stay on the right side of the sidewalk. Unfortunately, although these behaviors are instinctual to humans, they are hard to quantify and may vary depending on the person, culture, or region \cite{variance}. Therefore, they are difficult to reproduce artificially.

Although previous robotic navigation works have made advances in the building of approaches that respect social norms, many unsolved problems remain. The majority of the models work very well in the proposed environment, but get terrible results when evaluated in new environments \cite{deepmotion,rlblogpost}. 
In this work, we discuss this problem and possible solutions by analysing a previous study on adaptive convolutional encoders and Deep Reinforcement Learning (DRL) for navigation.









\section{RELATED WORKS}

\subsection{Convolutional Neural Network Encoder (CNN Encoder)}

Janderson et al. proposed a Deep Network-based encoder able to reduce the number of routes in path planning problems \cite{jds}. The input of this CNN Encoder is an image with the start point and the goal, as well many fixed and random obstacles. The model's label is the shortest path between the start point and the goal. The metrics employed were the number of iterations and time spent. It was found that the Encoder combined with conventional path planning algorithms significantly improved performance when compared to their isolated use, the average reduction in the execution time being 54.43\% \cite{jds2}. Such an improvement is important for robotic navigation in social situations, since in these scenarios robots need to constantly recalculate routes due to changes in the environment.




\subsection{Socially Aware Motion Collision Avoidance with Deep Reinforcement Learning}

Chen et al. developed a socially aware collision avoidance with deep reinforcement learning (SA-CADRL) model that aims to reproduce human observance of social norms \cite{sacadrl}. It improves upon previous works that primarily attempted to imitate human behavior but failed to generalize due to high variance in the feature values from different people. 

They posit that social transit norms are not, in fact, a set of predefined rules, but rather stem from time-efficient, reciprocal collision avoidance. That is, each agent produces a model of the other's behavior and reacts without explicitly communicating intention. Therefore, SA-CADRL includes both a reward function for efficient use of time and a reciprocity assumption. Furthermore, socially desirable behavior is introduced through a small bias that favors a decision over another, penalizing the unwanted action.

Although the original collision avoidance with deep reinforcement learning (CADRL) model could be scaled up to multiagent scenarios, it experienced a decline in performance because it was only trained on two agents \cite{cadrl}. This was also improved in SA-CADRL, which was trained with multiple agents.

This work also demonstrates the model's behavior on a fully autonomous robot, navigating in a pedestrian-rich environment at human walking speed.


\subsection{DeepMoTIon: Learning to Navigate Like Humans}

The DeepMoTIon: Learning to Navigate Like Humans (DeepMoTIon) model proposed by Hamandi et al. intends to mimic human navigation \cite{deepmotion}. It is trained on a pedestrian dataset of a real-world environment, with dense crowds and abrupt movements. The training approach is to replace humans in the data with the robot, which will survey the environment at every individual time step.

To account for natural variances in human movement, they propose a loss function that uses a Gaussian distribution to penalize the robot less for instances where it obtained results close to the ground truth. However, the standard deviation cannot be abstracted from the dataset, given that it does not repeat the same situation in varied samples. Therefore, it is a hyperparameter, tuned during training.

In this manner, DeepMoTIon addresses the issue pointed out by Chen et al., wherein algorithms that mimic human transit were not able to generalize \cite{sacadrl}. Furthermore, unlike SA-CADRL, it does not require manually inputting social norms and, therefore, does not need to be altered for differing cultural or regional scenarios.

This model outperformed all current benchmarks on path difference and proximity metrics, and all but one on the number of collisions.

\section{Analysis of the CNN Encoder and Deep Reinforcement Learning to Incremental Learning in Social Robotic Navigation}


The CNN Encoder was proposed combining the ability to extract characteristics from convolutional neural networks with the reconstruction of input data at a reduced level of Autoencoders complexity. Based on this model, we now have access to a solution that can be easily integrated as a module for preprocessing the inputs of search algorithms already common in the literature.

\begin{figure}[htbp]
\label{rl_fail}
\centerline{\includegraphics[width=.4\textwidth]{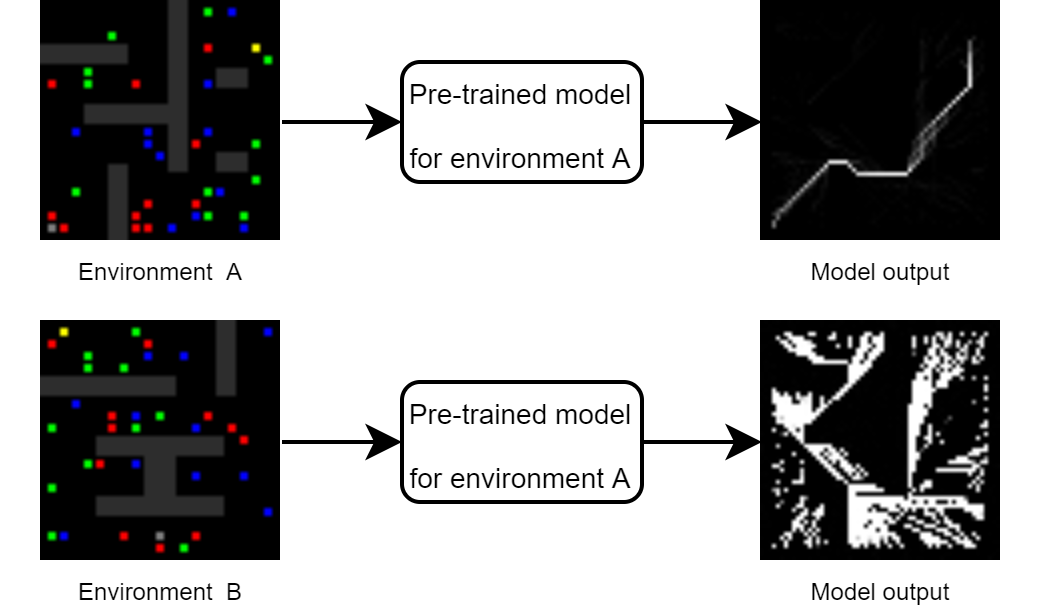}}
\caption{CNN Encoder generalization failed in different environments.}
\label{cnn_encoder}
\end{figure}

However, CNN Encoder has some limitations that prevent it from being used in social navigation tasks. This model has no response to the changes in the environment being applied, as is illustrated in Figure \ref{cnn_encoder}. In this experiment, we used our database, also used to train the CNN Encoder, which contains five scenarios, each with a total of 10000 scenes (60 x 60 pixels, RGB images). A label (60 x 60 pixels, grayscale image) accompanies each of them, containing the shortest path to solve the problem.


\vspace{-1em}
\vspace{-1em}

\begin{figure}[htbp]
\centerline{\includegraphics[width=.3\textwidth]{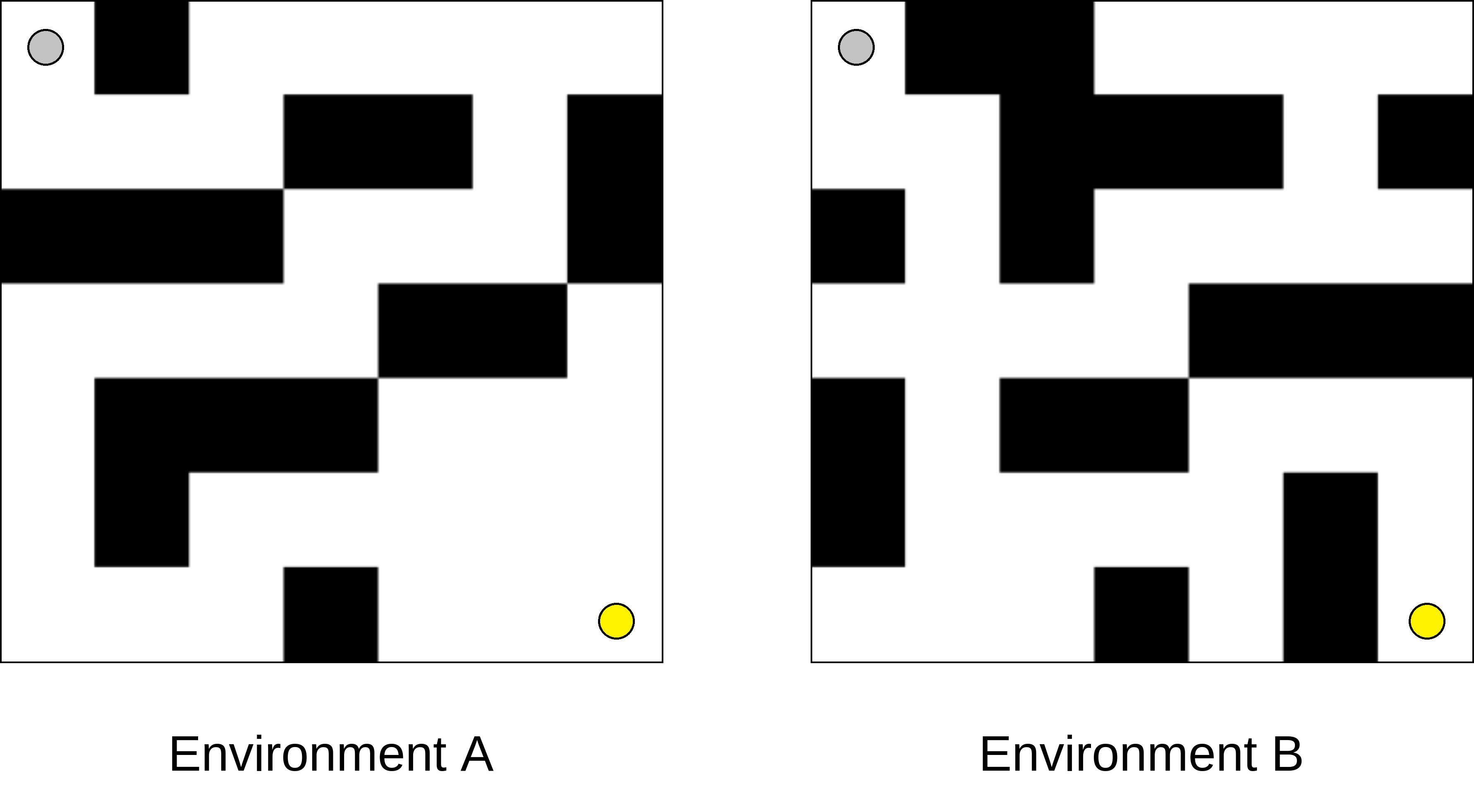}}
\caption{DRL generalization failed with environmental changes.}
\label{ambientes}
\end{figure}

While some studies demonstrate that DRL can be a viable solution for social navigation problem, they do not account for the diversity in environments \cite{sacadrl,deepmotion}. With unpredictable scenarios, the number of situations becomes infinite. Thus, during our tests on the two environments that we created, it was noticed that DRL algorithms do not converge with continuous changes in the original environment. Even when only a few changes are made, DRL does not have predictable results. We conducted a simple experiment to detect possible limitations of DRL models. Figure \ref{ambientes} shows this experiment.
\vspace{-1em}

\vspace{-1.5em}
\begin{table}[htbp]
\centering
\begin{tabular}{|l|c|c|c|}
\hline
Environment & \multicolumn{1}{l|}{Win Rate} & \multicolumn{1}{l|}{Run Time} & \multicolumn{1}{l|}{Episodes} \\ \hline
A      & 98\%             & 113.4 s            & 50              \\ \hline
B      & 6\%              & 318.7 s            & 50              \\ \hline
\end{tabular}
\caption{Results of the algorithms in the environments presented.}
\label{table_results}
\end{table}
\vspace{-2em}


The data in table \ref{table_results} demonstrates that DRL does not fit as a reasonable answer to the problem, since learning is not guaranteed for situations different from the ones it knows. Additionally, the computational cost of training these kinds of models is very high \cite{rlblogpost}.

Considering the CNN Encoder's potential to reduce dimensionality and the improvement to social navigation made by DRL, a combination of these approaches to create an Incremental Learning Model can overcome the lack of generalization and the high cost to train DRL models.

\bibliographystyle{unsrt}
\bibliography{references}

\end{document}